\documentclass[letterpaper, 10 pt, conference]{ieeeconf}  

\IEEEoverridecommandlockouts  

\usepackage{times}

\usepackage{cite}
\usepackage{multicol}
\usepackage{graphics} 
\usepackage{amsmath} 
\usepackage{amssymb}  
\usepackage{algorithm}
\usepackage{algpseudocode}
\usepackage{subcaption}
\usepackage[dvipsnames]{xcolor}
\usepackage{hyperref}
\hypersetup{
    colorlinks=true,
    linkcolor=black,    
    urlcolor=black,
    citecolor=black
    }

\usepackage{epsfig} 
\usepackage{times} 
\usepackage{amsmath} 
\usepackage{amssymb}  

\usepackage{algpseudocode}

\title{\Large \bf Adaptive Visual Imitation Learning for Robotic Assisted Feeding \\Across Varied Bowl Configurations and Food Types}

\author{Rui Liu, Amisha Bhaskar, Pratap Tokekar   
\thanks{All authors are from the Department of Computer Science, University of Maryland, College Park, MD 20742, USA. Email: \tt\small \{ruiliu, amishab, tokekar\}@umd.edu}}

\begin{document}

\maketitle

\begin{abstract}
In this study, we introduce a novel visual imitation network with a spatial attention module for robotic assisted feeding (RAF). The goal is to acquire (i.e., scoop) food items from a bowl. However, achieving robust and adaptive food manipulation is particularly challenging. To deal with this, we propose a framework that integrates visual perception with imitation learning to enable the robot to handle diverse scenarios during scooping. Our approach, named AVIL (adaptive visual imitation learning), exhibits adaptability and robustness across different bowl configurations in terms of material, size, and position, as well as diverse food types including granular, semi-solid, and liquid, even in the presence of distractors. We validate the effectiveness of our approach by conducting experiments on a real robot. We also compare its performance with a baseline. The results demonstrate improvement over the baseline across all scenarios, with an enhancement of up to 2.5 times in terms of a success metric. Notably, our model, trained solely on data from a transparent glass bowl containing granular cereals, showcases generalization ability when tested zero-shot on other bowl configurations with different types of food.
\end{abstract}

\section{Introduction}
Robotic Assisted Feeding (RAF) refers to the use of robotic systems to assist individuals with feeding disabilities or impairments in their daily lives. Researchers have explored various approaches to enhance the autonomy and efficiency of meal assistance for this demographic. Some have focused on integrating sensors, particularly visual and haptic ones \cite{bhattacharjee2019towards, goldau2019autonomous, sundaresan2022learning, gordon2021leveraging}, to interact with food items and dynamically adjust feeding strategies based on sensor feedback. Bhattacharjee et al. \cite{bhattacharjee2019towards} investigated the role of haptics in fork-based food manipulation. Goldau et al. \cite{goldau2019autonomous} presented an autonomous multi-sensory robotic assistant for a drinking task. Sundaresan et al. \cite{sundaresan2022learning} proposed a framework for sensing visuo-haptic properties of food items and subsequently skewering them. Gordon et al. \cite{gordon2021leveraging} leveraged the haptic context collected during and after manipulation to learn food properties and adapt to previously unseen food. 

Others have explored the development of RAF systems equipped with computer vision algorithms \cite{gallenberger2019transfer, feng2019robot, gordon2020adaptive, grannen2022learning, tai2023scone} to recognize food items and provide precise and timely feeding assistance based on user preferences. Gallenberger et al. \cite{gallenberger2019transfer} developed a robotic feeding system employing skewering and transfer primitives for autonomous feeding. Feng et al. \cite{feng2019robot} introduced a bite acquisition framework, SPANet, capable of generalizing skewering strategies across various food items on a plate. Gordon et al. \cite{gordon2020adaptive} adapted SPANet and used contextual bandit to handle unfamiliar food items and decide the best primitive selection strategy based
on real-time interactions. Grannen et al. \cite{grannen2022learning} focused on learning bimanual scooping policies for food acquisition, incorporating a pusher to stabilize food items during the scooping process. Tai et al. \cite{tai2023scone} presented a food-scooping robot learning framework that leverages representations gained from active perception to facilitate food scooping.

\begin{figure}
    \centering
    \includegraphics[width=\linewidth]{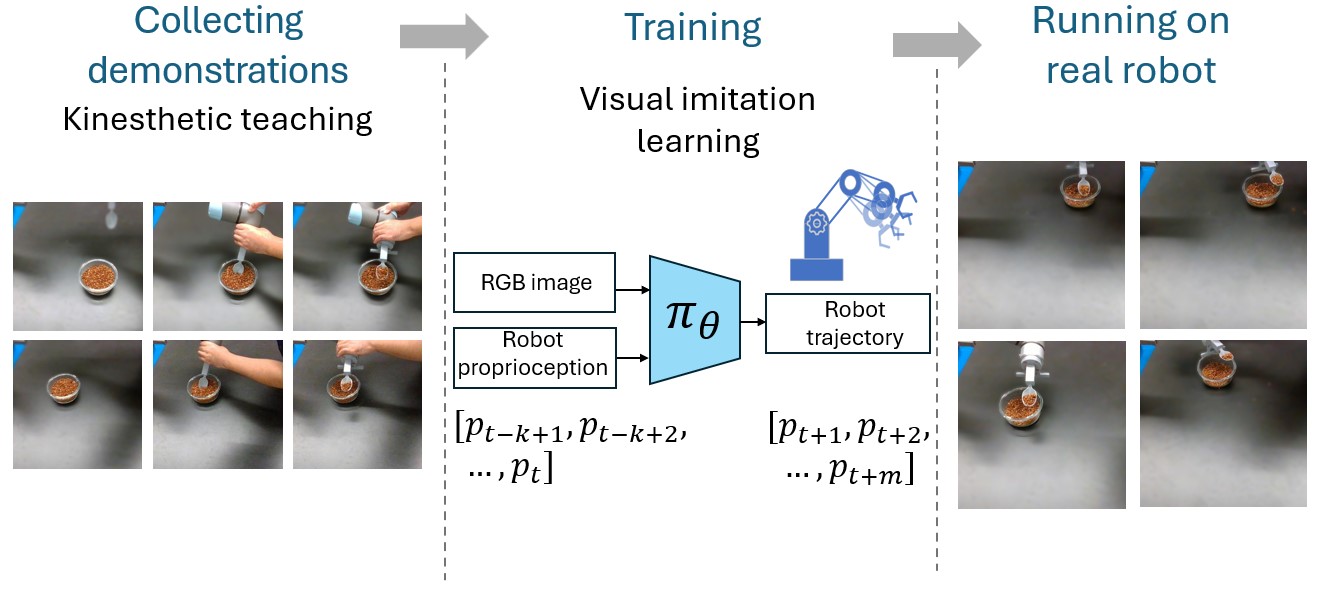}
    \caption{Learning pipeline diagram of our approach (\textbf{AVIL}) for spoon scooping in RAF.}
    \label{fig:pipeline}
\end{figure}

While prior work has achieved significant progress, certain limitations persist. Previous research on RAF has mainly focused on learning individual low-level parameterized primitives for food manipulation, whether fork-based skewering \cite{gallenberger2019transfer, feng2019robot} or spoon-based scooping \cite{tai2023scone}. These policies suffer from limitations such as handcrafted trajectories or limited adaptability to different food types and container configurations, including variations in position, size, and material. Achieving robust and adaptive food manipulation is particularly challenging when faced with variability in container configurations and food types. For instance, Tai et al. \cite{tai2023scone} solely considered granular food like beans and rice, neglecting variations in bowl materials, sizes, positions, and other food types including semi-solid and liquid.

In contrast, humans effortlessly adapt to different container configurations and food types while eating, exhibiting behavior that is inherently natural. Inspired by human eating behavior, we pose a question: \textit{Can robots learn to acquire food of various types from human demonstrations for assisted feeding?}

In this study, we focus on spoon scooping, an essential aspect of RAF, aiming to address this question and effectively scoop food from a bowl. Toward this objective, we developed a novel visual imitation network to overcome the aforementioned limitations and achieved adaptive scooping across varied bowl configurations and food types. The network incorporates a spatial attention module, illustrated in Figure \ref{fig:vilnet}, which dynamically assigns weights to different spatial locations in the input image, enabling the network to only focus on the area of interest. We name our approach \textbf{AVIL}, which stands for Adaptive Visual Imitation Learning. 


To validate our approach, we tested it on a real robot and compared its performance with a baseline represented by a handcrafted scooping motion. The results demonstrate that our model outperforms the baseline across all varied bowl configurations and food types. Notably, we trained the model solely with data collected from granular cereals in a transparent glass bowl. Despite this, the model exhibited effectiveness when tested zero-shot with plastic bowls of different sizes, as well as with other food types such as semi-solid jelly and liquid water. 


Moreover, we assessed the model's robustness by subjecting it to tests with distractors on the table. Even in the presence of distractors, the model maintained its performance, showcasing its robustness and resilience.

To summarize, the key contributions of our work as follows:
\begin{itemize}
    \item We introduce a novel visual imitation network with a spatial attention module for spoon scooping in RAF.  
    \item Our approach demonstrates adaptability and robustness, performing effectively with varied bowl configurations, food types, and in the presence of distractors. This overcomes drawbacks of previous work with limited adaptability to different container configurations and food types. 
    \item We conduct comprehensive experiments on a real robot. The results demonstrate clear improvement over baseline, validating the efficacy of our model. 
\end{itemize}


\section{Related Work} \label{sec:related}

In this section, we review related work on RAF and different techniques applicable to RAF, such as imitation learning and vision-based robotic manipulation.

\subsection{Robot Assisted Feeding}
In the realm of robot-assisted feeding, researchers have explored various approaches to enhance the autonomy and efficiency of meal assistance for individuals with disabilities or impairments. Park et al. \cite{park2016towards} presented a proof-of-concept robotic system for assistive feeding with a mobile manipulator. Bhattacharjee et al. \cite{bhattacharjee2019towards} explored the role of haptics in fork-based food manipulation. Gallenberger et al. \cite{gallenberger2019transfer} designed both skewering and transfer primitives and developed a robotic feeding system that uses these manipulation primitives to feed people autonomously. Goldau et al. \cite{goldau2019autonomous} presented an autonomous multi-sensory robotic assistant for a drinking task. One notable line of investigation involves the development of robotic systems equipped with computer vision algorithms, enabling robots to adapt to different food types and user preferences. Feng et al. \cite{feng2019robot} developed SPANet which displays proficient capability in handling a diverse range of food items by mapping food image observations to actions. Gordon et al. \cite{gordon2020adaptive} adapted SPANet and used contextual bandit to handle unfamiliar food items and decide the best primitive selection strategy based on real-time interactions. Sundaresan et al. \cite{sundaresan2022learning} leveraged visual and haptic observations to learn skewering strategies with a fork to acquire food items.

Previous work on RAF has mainly focused on learning specific actions for handling food. However, achieving robust and adaptive food manipulation is particularly challenging. We propose a framework that integrates visual perception with imitation learning to enable the robot to handle different bowl configurations and food types during scooping. Moreover, our approach showcases robustness even when there are distractions. This allows the robot to handle real-world feeding scenarios more effectively.



\subsection{Imitation Learning}
Imitation learning (IL) enables robots to acquire new skills and knowledge by observing demonstrations provided by experts. The robot's goal is to mimic the expert's behavior as closely as possible. It has gained significant attraction in robotics and autonomous systems especially when the task cannot be easily programmed, but can be demonstrated. Hussein et al. \cite{hussein2017imitation} surveyed imitation learning methods and present design options in different steps of the learning process. Fang et al. \cite{fang2019survey} also provided a survey of imitation learning of robotic manipulation and explore the future research trend from three main aspects: demonstration, representation and learning algorithms. Johns \cite{johns2021coarse} proposed a coarse-to-fine imitation learning approach for robot manipulation tasks. Young et al. \cite{young2021visual} presented an alternate interface for imitation learning that simplifies the data collection process for learning complex robot manipulation tasks. 

\subsection{Vision-Based Robotic Manipulation}
In the field of vision-based manipulation, researchers have been focusing on methods to enable robots to understand and interact with their surroundings using visual cues. Kalashnikov et al. \cite{kalashnikov2018scalable} studied the problem of learning vision-based dynamic manipulation skills using a scalable reinforcement learning approach. Ali et al. \cite{ali2018vision} presented a multi-stage process of the development of a vision-based object sorting robot manipulator for industrial applications. Julian et al. \cite{julian2020efficient} demonstrated how to adapt vision-based robotic manipulation policies to new variations by fine-tuning via off-policy reinforcement learning. Cong et al. \cite{cong2021comprehensive} reviewed recent progress in the 3-D vision for robot manipulation, such as 3-D object detection, 6-DOF pose estimation, grasp estimation, and motion planning. 

\begin{figure*}[hbt]
    \centering
    \includegraphics[width=\textwidth]{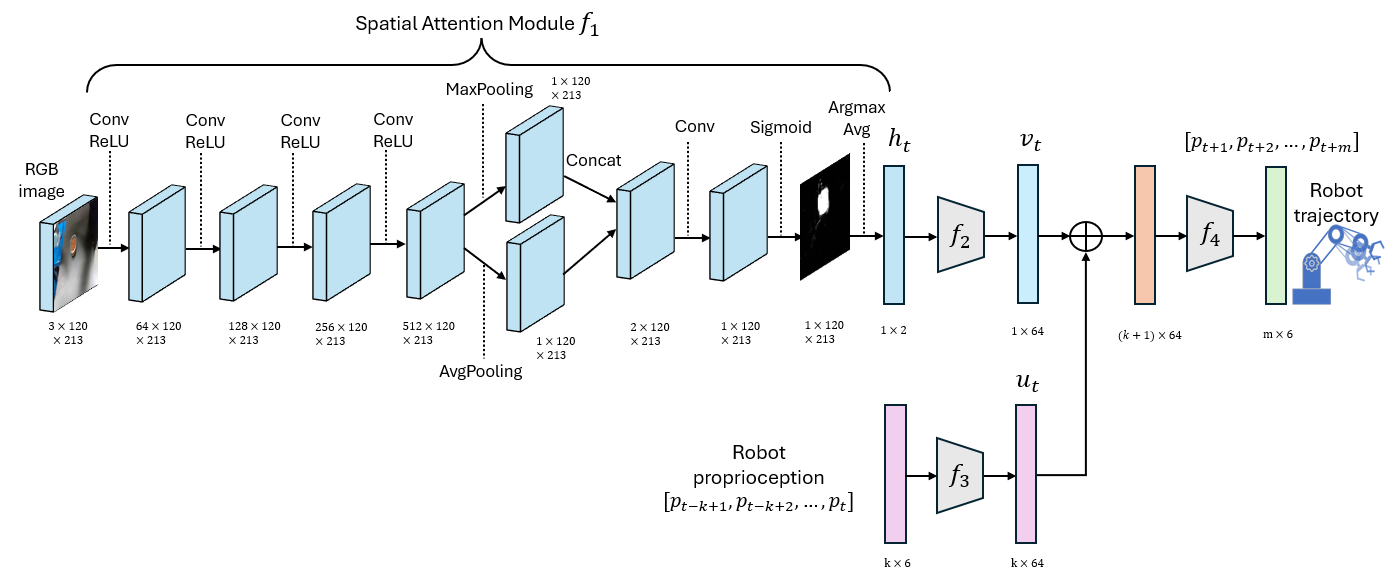}
    \caption{Proposed visual imitation network.}
    \label{fig:vilnet}
\end{figure*}

\section{Approach} \label{sec:approach}
In this study, we focus on the learning of visuomotor policies for RAF, specifically involving spoon scooping for varied bowl configurations and food types. Employing visual imitation learning, we build a robust framework that integrates imitation learning to directly map input observations, including RGB images and robot proprioception (joint positions), to corresponding robot control actions. The learned policy is adaptive to variations in bowl position, size, material, and food types, making it well-suited for RAF tasks. Figure \ref{fig:pipeline} illustrates our learning pipeline. This process involves collecting human expert demonstrations, training the model using our visual imitation network, and then deploying the learned policy on a real robot.


\subsection{Preliminaries}
\subsubsection{Observation and Action Space}
In our visuomotor policy learning system of spoon scooping for RAF, the input observation space $\mathcal{O}_t = (\mathcal{I}_t, p_t)$, where $\mathcal{I}_t \in \mathbb{R}^{3\times H \times W}$ represents the RGB images captured from a static environment camera, and $p_t \in \mathbb{R}^6$ denotes the robot proprioception, representing the 6D joint positions. The state $s_{t-k:t} = (\mathcal{I}_t, p_{t-k+1}, p_{t-k+2}, \ldots, p_t)$ involves the RGB image of the current timestep $t$ and a sequence of last $k$ steps of joint positions. The action $a_{t:t+m} = (p_{t+1}, p_{t+2}, \ldots, p_{t+m}) \in \mathbb{R}^{m \times 6}$ involves the predicted joint positions for the next $m$ steps. 

\subsubsection{Visuomotor Policy Learning}
In the domain of visuomotor policy learning, the primary objective is to learn a policy $\pi: \mathcal{S} \rightarrow \mathcal{A} $ that guides a robot's actions in manipulating objects. In this study, we choose Behavior Cloing (BC) as the imitation learning algorithm. We formulate the policy learning problem as a supervised learning task, aiming to learn a parameterized policy $\pi_\theta$ with the following objective function:
\begin{equation}\label{eq:obj}
    \theta = \arg \min_\theta \mathbb{E}_{(s_{t-k:t}, a_{t:t+m}^*) \sim \mathcal{D}} [\mathcal{L}(\pi_\theta(s_{t-k:t}), a_{t:t+m}^*)], 
\end{equation}
where $\theta$ is the parameters of the policy, $\pi_\theta(s_{t-k:t})$ is the predicted actions for the state $s_{t-k:t}$, and $a_{t:t+m}^*$ is the expert action. $\mathcal{L}$ is the loss function, which is the mean squared error loss in our case. 

To initiate the visuomotor policy learning process, we first collect a set of demonstrations as the training data. Our dataset $\mathcal{D}=\{(s_{t-k:t}, a_{t:t+m}^*)\}_{i=1}^N$ consists of $N$ robot trajectories obtained through kinesthetic teaching, where each trajectory $i$ comprises pairs of states and actions.

After completing the preliminaries and formulating the visuomotor policy learning problem, we proceed to tackle the problem and learn the policy using our visual imitation network.

\subsection{Visual Imitation Policy Network} \label{sec:vinet}
In this section, we introduce the main component of our paper, the visual imitation policy network, designed to learn visuomotor policies for scooping various food items. The network is crucial for mapping visual observations and robot proprioception into robot actions. The network's objective is to minimize the discrepancy between the predicted robot actions $\pi_\theta(\cdot)$ and the demonstrated actions throughout the learning process. 

To enhance learning from the historical data, our policy network takes the RGB image of current timestep $t$ and a sequence of last $k$ steps of robot proprioception (joint positions), $(\mathcal{I}_{t}, p_{t-k+1}, p_{t-k+2}, \ldots, p_{t})$, as input. The network's outputs consist of the predicted joint positions for the subsequent $m$ steps, $a_{t:t+m} = \pi_\theta(s_{t-k:t})$, which is a practical innovation. This approach, unlike solely predicting a single step in previous work \cite{zhang2018deep}, offers the advantage of mitigating error accumulation over time and providing a more detailed view on future trajectories. During inference on a real robot, we adopt the initial step from the predicted $m$ future steps for execution, drawing inspiration from Model Predictive Control (MPC). Then we observe and update the state $s_{t-k+1:t+1} = (\mathcal{I}_{t+1}, p_{t-k+2}, \ldots, p_t, p_{t+1})$. 




The architecture of our visual imitation network, outlined in Figure \ref{fig:vilnet}, incorporates several key components: a spatial attention module, a visual attention embedding module, a robot proprioception embedding module, and a control action module. The output layer generates predicted joint positions for the next $m$ steps, which are subsequently compared with expert actions during the training process. Notably, our proposed network incorporates a spatial attention module. This module enables the model to focus on crucial areas within the image, facilitating improved adaptation for different bowl configurations and food types for our RAF system.

\paragraph{Spatial attention module} \label{p:attention}
Spatial attention mechanism \cite{zhu2019empirical, woo2018cbam, chen2021attention} plays a crucial role in enhancing the model's ability to focus on relevant regions within the visual input. Through the incorporation of a spatial attention module, our model dynamically weights different spatial locations in the input image. This allows the robot to selectively attend to specific areas of interest, such as the food source or the targeted bowl in our case, contributing to the development of a more refined and context-aware RAF system. The significance of spatial attention extends beyond our system of robot-assisted feeding, proving valuable in various other robot learning tasks \cite{shridhar2022cliport, zeng2021transporter, seita2021learning}.

We depict the architecture of the spatial attention module in Figure \ref{fig:vilnet}. The module comprises four convolution layers for image feature extraction, each followed by a ReLU activation \cite{nair2010rectified}. Each layer employs a filter size of 3, stride 1, and padding 1. In this way, we can maintain the spatial dimensions of the feature map for an input image while increasing its depth.

Drawing inspiration from the convolutional block attention module in \cite{woo2018cbam}, we apply channel-wise max pooling and average pooling to the intermediate feature map, resulting in two one-dimensional feature maps. Then we concatenate these two feature maps along the channel, and pass through a convolution layer and a sigmoid layer, we obtain the spatial attention map. Notably, different from the approach in \cite{woo2018cbam}, we introduce an auxiliary binary cross-entropy loss to guide the learning of the spatial attention module.

\paragraph{Visual attention embedding module}
A linear projection layer that projects the visual attention feature obtained from the spatial attention module into higher dimensions. 

\paragraph{Robot proprioception embedding module}
A linear projection layer that projects the robot proprioception (joint angles) into higher dimensions. 

\paragraph{Control action module}
A Multi-Layer Perceptron (MLP) with several fully connected layers for mapping features to robot actions.

As mentioned before, we can decompose our visual imitation network $\pi_\theta$ into four modules. Mathematically, $\pi_\theta = (f_1, f_2, f_3, f_4)$, where $f_1$ is the spatial attention module with parameters $\theta_1$, $f_2$ is the visual attention embedding module with parameters $\theta_2$, $f_3$ is the robot proprioception embedding module with parameters $\theta_3$, and $f_4$ is the control action module with parameters $\theta_4$.
 \begin{align}
     h_t &= f_1(\mathcal{I}_{t}; \theta_1), \\
     v_t &= f_2(h_t; \theta_2), \\
     u_t &= f_3(p_{t-k+1}, p_{t-k+2}, \ldots, p_{t}; \theta_3), \\
     a_{t:t+m} &= f_4(v_t, u_t; \theta_4).
 \end{align}

For our proposed visual imitation network, given the RGB image $\mathcal{I}_t$, it initially undergoes the spatial attention module $f_1$, producing a one-channel attention map. By performing an argmax operation on this map, we derive the coordinates representing the area of interest in the image. A subsequent averaging operation allows us to determine the 2D visual attention feature $h_t = (x_t, y_t)$, essentially representing the centroid of this region. Then we project $h_t$ into higher 64 dimensions via the linear embedding ${f_2}$. Simultaneously, the last $k$ steps of 6D robot joint positions are also projected into higher 64 dimensions via the linear embedding ${f_3}$. By projecting data into a higher-dimensional space, our model gains more capacity to represent complex functions. This increased expressiveness allows the model to capture non-linear relationships that might not be discernible in lower dimensions. We concatenate the resulting visual feature vector $v_t$ and embedded robot proprioception vector $u_t$ and proceed through the control action module $f_4$, ultimately produce the robot actions $a_{t:t+m}$.


\begin{figure*}[hbt!]
    \centering
    \includegraphics[width=0.9\textwidth]{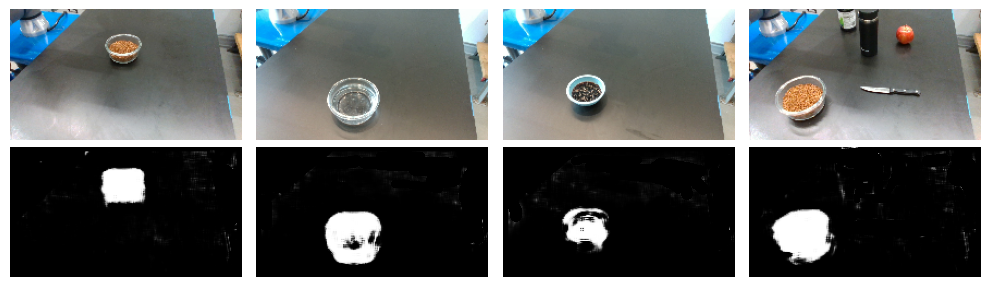}
    \caption{Qualitative results of images with various bowl configurations, positions, food types, and with distractors on the table along with their corresponding spatial attention maps.}
    \label{fig:att_map}
\end{figure*}

\subsection{Training the Network} \label{sec:train}
We initiate the training process by focusing on the spatial attention module. After collecting the demonstration dataset $\mathcal{D}$, we proceed to label the bowl in images with bounding boxes using LabelImg \cite{tzutalin2015labelimg}, creating masks to isolate the relevant information. In total, we generate 1304 masks for different bowl positions. These masks are then compared with the spatial attention map generated in Section \ref{sec:vinet}, employing binary cross-entropy loss. Subsequently, we freeze the weights of the spatial attention module. Following this step, we proceed to train the entire visual imitation network using mean squared loss. To balance prediction accuracy and complexity, we set $k=4$ and $m=2$, which involves stacking the historical data of the last 4 steps for the current timestep and predicting actions for the next 2 steps.

For both the spatial attention module and the entire visual imitation network, we conduct training on an Nvidia RTX 3090 GPU for 200 epochs. We utilize a learning rate of $10^{-4}$ and a batch size of 8, employing the Adam optimizer \cite{kingma2014adam} along with a cosine learning rate scheduler \cite{loshchilov2017decoupled}.

We present some qualitative results that highlight the effectiveness of our spatial attention module and its robustness against distractors after training. We display images depicting various bowl configurations, positions, and food types along with their corresponding spatial attention maps in Figure \ref{fig:att_map}. Specifically, we show a transparent glass bowl containing cereals, followed by its contents shifting to water at a different position. Subsequently, we depict a plastic bowl containing jelly. Additionally, we introduce distractor items such as a water bottle, an apple, a jelly jar, and a knife onto the table to simulate a realistic kitchen environment. The spatial attention module effectively focuses on the bowl area, as evidenced by the corresponding attention map. This allows the visual imitation network to accurately scoop the desired food despite the presence of distractors. Notably, our spatial attention module is much smaller, requires less computation, and is faster to train compared to pretrained object detectors like RetinaNet \cite{lin2017focal}, Yolo \cite{redmon2016you}, and Mask-RCNN \cite{he2017mask}, yet achieves effective performance in extracting visual features for our RAF application.

\section{Experiments} \label{sec:exp}
We begin by describing the experimental setup. Following this, we discuss the data collection procedure and the baseline for comparison. Subsequently, we present and analyze the experimental results.

\subsection{Experimental Setup} \label{sec:exp_setup}
\begin{figure}[ht!]
    \centering
    \includegraphics[width=0.9\linewidth]{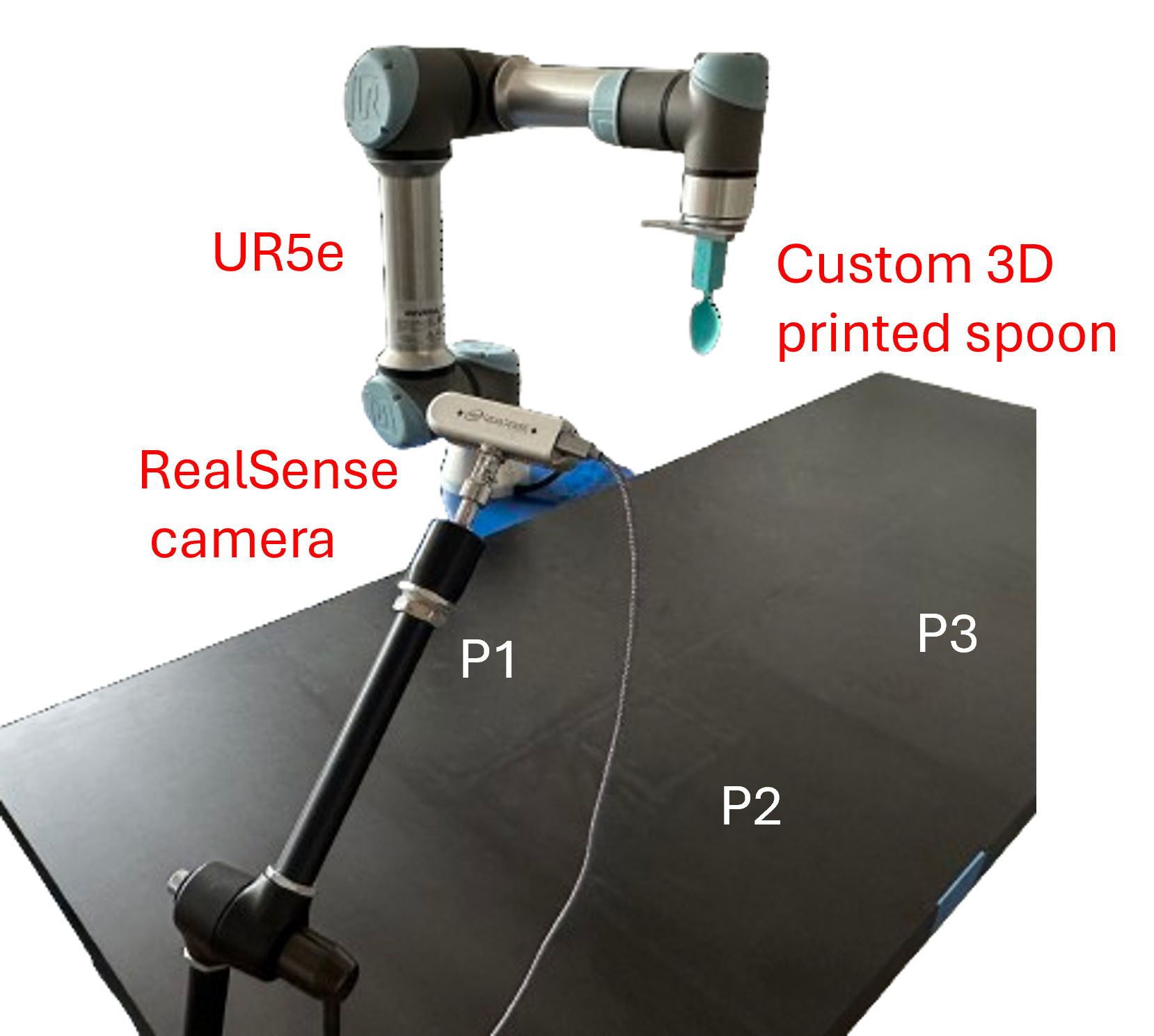}
    \caption{The experimental setup, which includes a UR5e robot arm, a custom-designed spoon attachment, and a stationary RealSense camera. P1, P2, P3 denote different bowl positions on the table.}
    \label{fig:setup}
\end{figure}

\begin{figure}[ht!]
     \centering
     \begin{subfigure}[b]{0.32\linewidth}
         \centering
         \includegraphics[width=\linewidth]{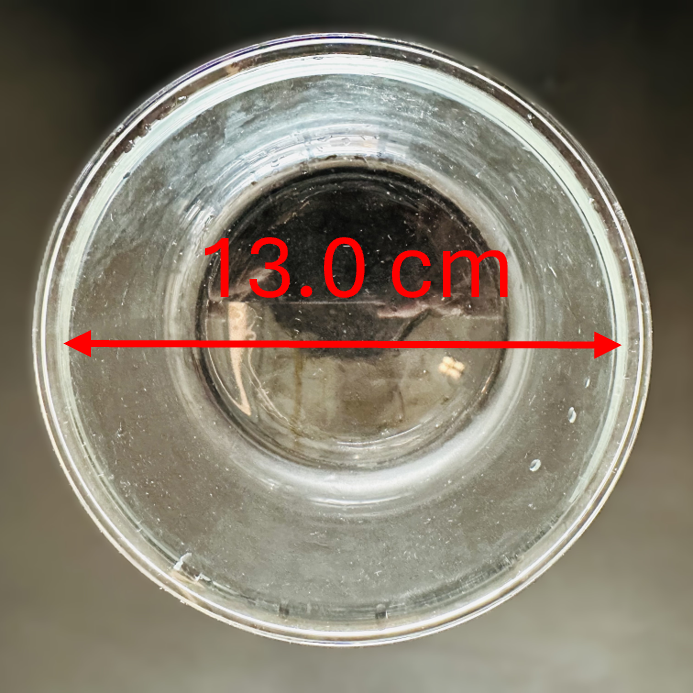}
         \caption{TG bowl}
         \label{fig:g_bowl}
     \end{subfigure}
     \begin{subfigure}[b]{0.32\linewidth}
         \centering
         \includegraphics[width=\linewidth]{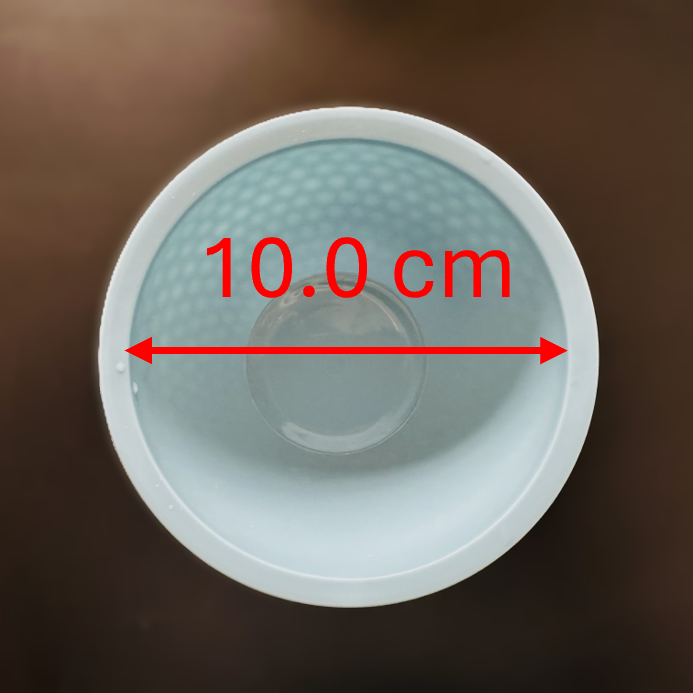}
         \caption{PS bowl}
         \label{fig:ps_bowl}
     \end{subfigure}
     \vfill
     \begin{subfigure}[b]{0.32\linewidth}
         \centering
         \includegraphics[width=\linewidth]{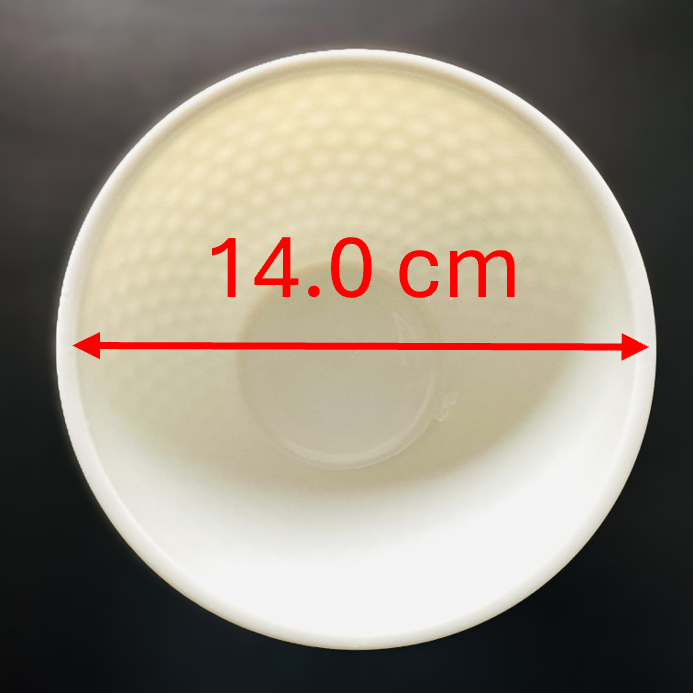}
         \caption{PM bowl}
         \label{fig:pm_bowl}
     \end{subfigure}
     \begin{subfigure}[b]{0.32\linewidth}
         \centering
         \includegraphics[width=\linewidth]{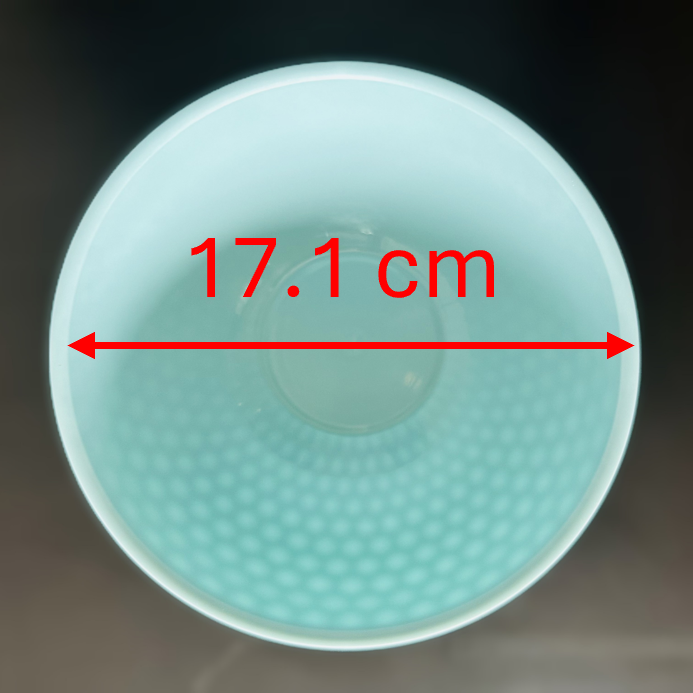}
         \caption{PL bowl}
         \label{fig:pl_bowl}
     \end{subfigure}
        \caption{Different bowl configurations exhibit variations in material, size, and color. TG denotes transparent glass, PS denotes plastic small, PM denotes plastic medium, and PL denotes plastic large.}
        \label{fig:bowls}
\end{figure}

\begin{figure}[ht!]
     \centering
     \begin{subfigure}[b]{0.32\linewidth}
         \centering
         \includegraphics[width=\linewidth]{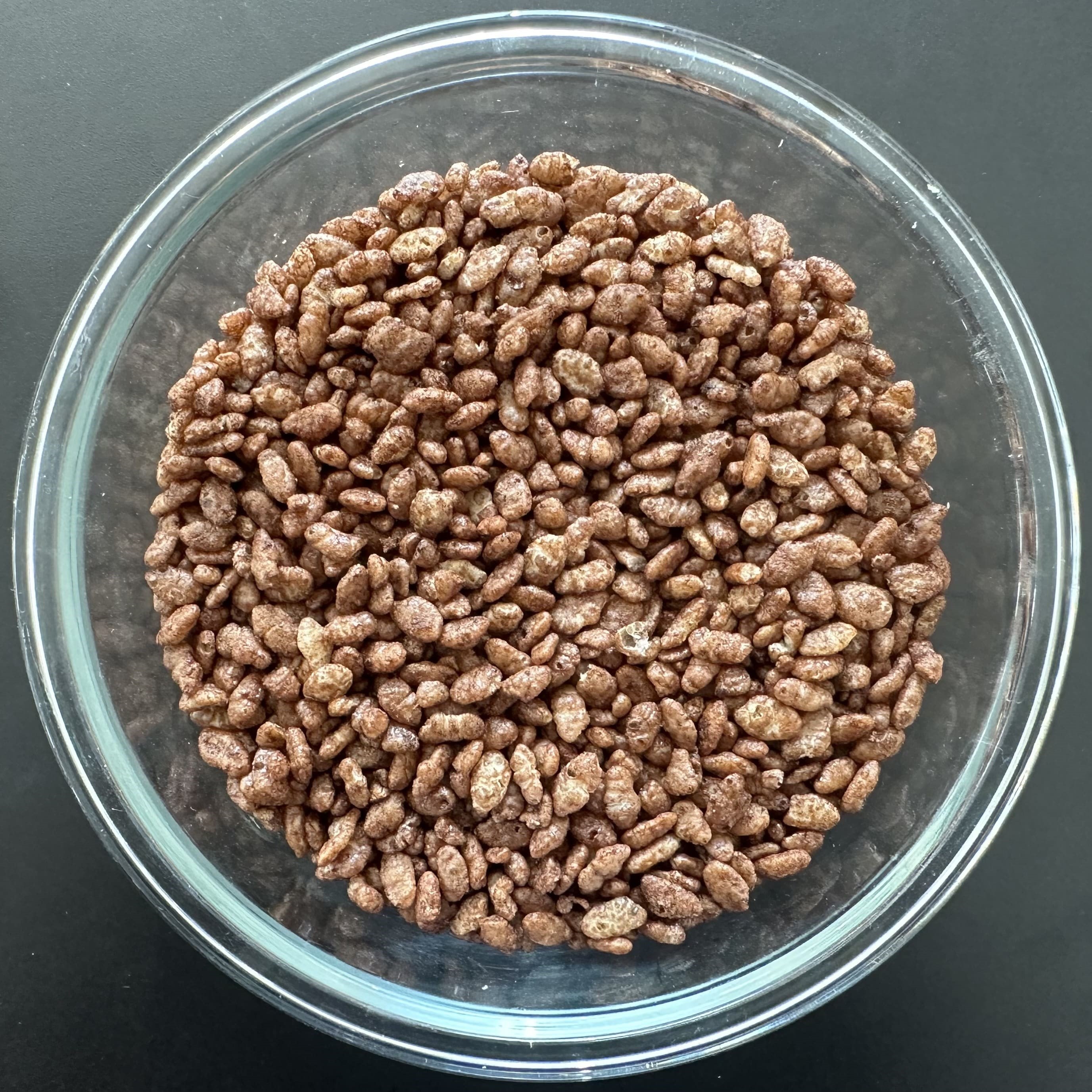}
         \caption{Granular (cereals)}
         \label{fig:cereals}
     \end{subfigure}
     \begin{subfigure}[b]{0.32\linewidth}
         \centering
         \includegraphics[width=\linewidth]{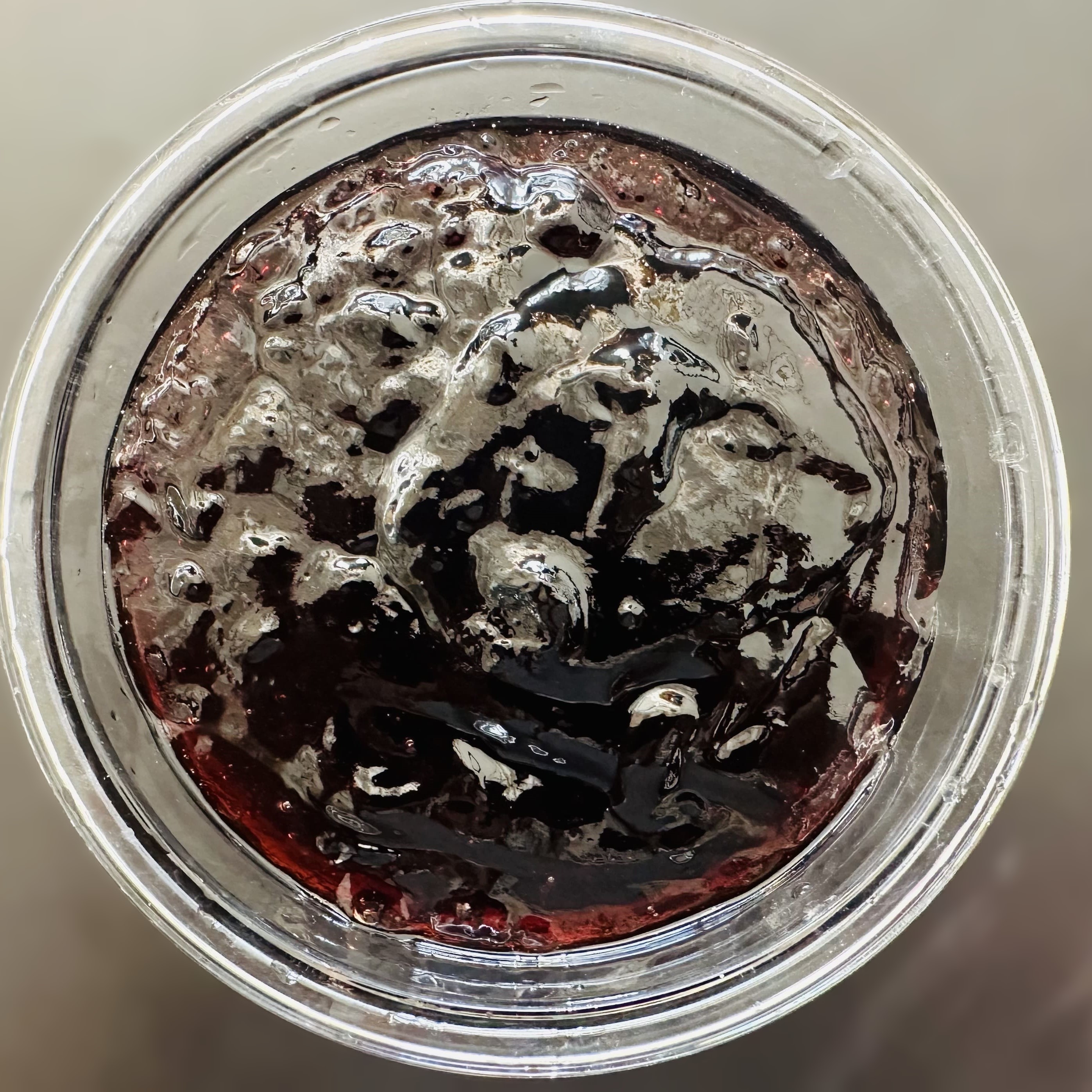}
         \caption{Semi-solid (jelly)}
         \label{fig:jelly}
     \end{subfigure}
     \begin{subfigure}[b]{0.32\linewidth}
         \centering
         \includegraphics[width=\linewidth]{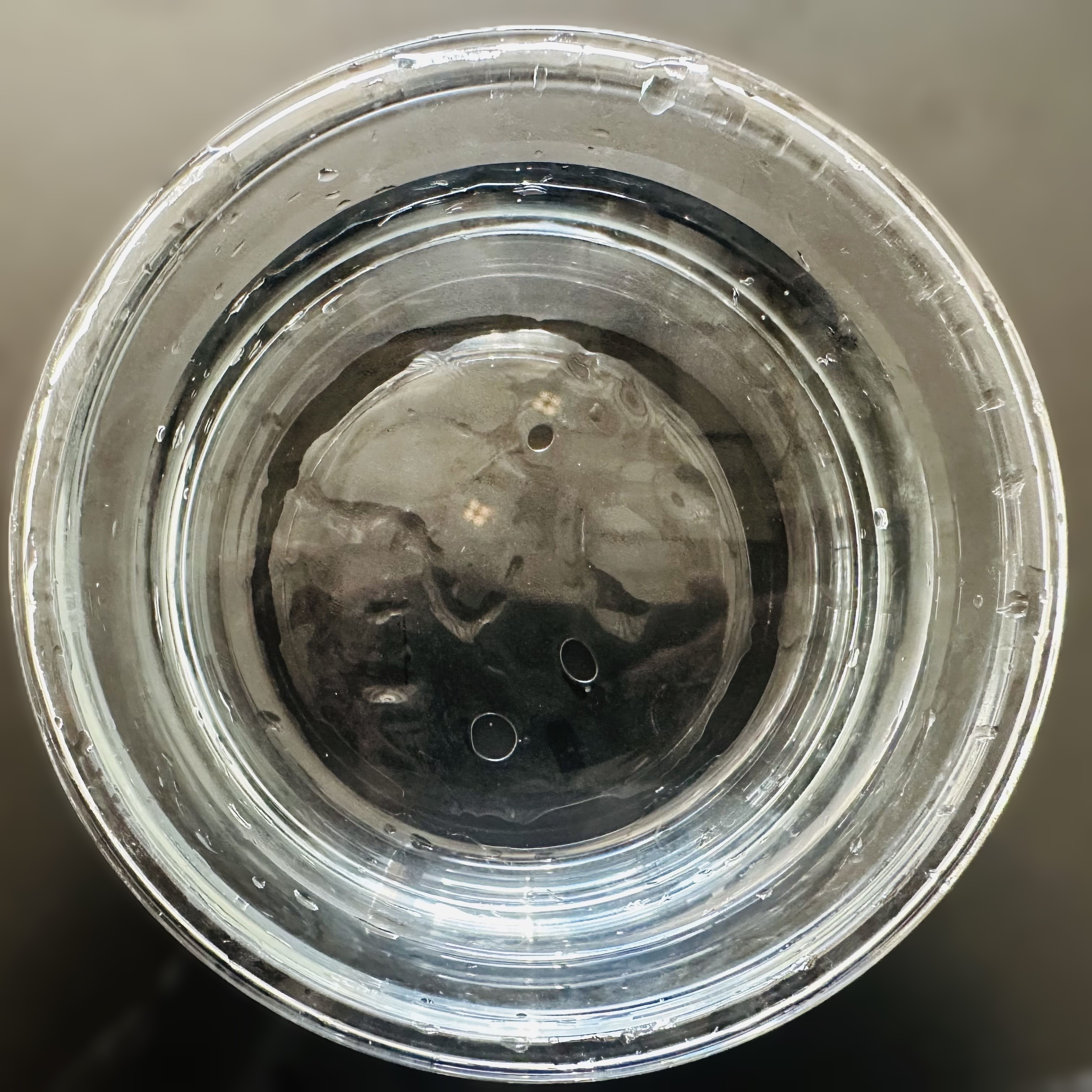}
         \caption{Liquid (water)}
         \label{fig:water}
     \end{subfigure}
        \caption{Different food types include granular cereals, semi-solid jelly and liquid water.}
        \label{fig:food}
\end{figure}

We illustrate the experimental hardware setup in Figure \ref{fig:setup}, which consists of several components: a UR5e robot arm, a custom-designed spoon attachment, and a stationary RealSense camera. The spoon is affixed to the arm, with a length measuring $10.0\ cm$. The RealSense camera is mounted on the table within the environment. During experiments, we explore varied bowl configurations across material, size, position, as well as different food types including granular cereals, liquid water, and semi-solid jelly, as shown in Figure \ref{fig:bowls} and Figure \ref{fig:food}. Bowl configurations encompass a transparent glass (TG) bowl, a small plastic (PS) bowl with a light blue color, a medium plastic (PM) bowl with a white color, and a large plastic (PL) bowl with a cyan color, each with differing diameters. Additionally, we conduct tests at different positions (P1, P2, P3) on the table, with their relative positions illustrated in Figure \ref{fig:setup}. It is important to note that the positions of P1, P2, and P3 are not precisely measured and fixed; instead, for each trial, any food spillage prompts us to clean and adjust the bowl position slightly. This variability in the testing environment allows us to evaluate the system's robustness under different conditions compared to the training phase.

\begin{figure*}[ht!]
    \centering
    \includegraphics[width=0.9\textwidth]{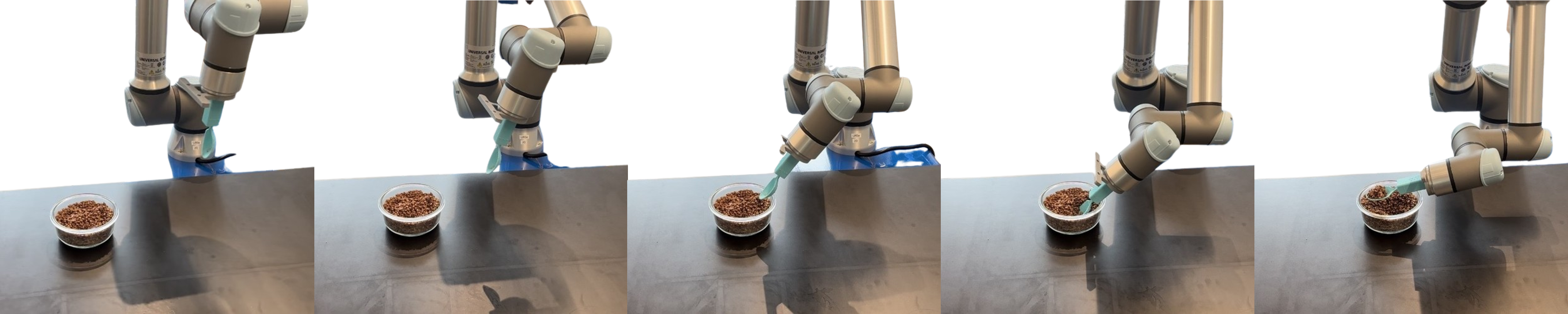}
    \caption{Example of sequential motion of UR5e successfully scooping granular cereals with \textbf{AVIL}.}
    \label{fig:seq}
\end{figure*}

\begin{figure}[ht!]
    \centering
    \includegraphics[width=0.8\linewidth]{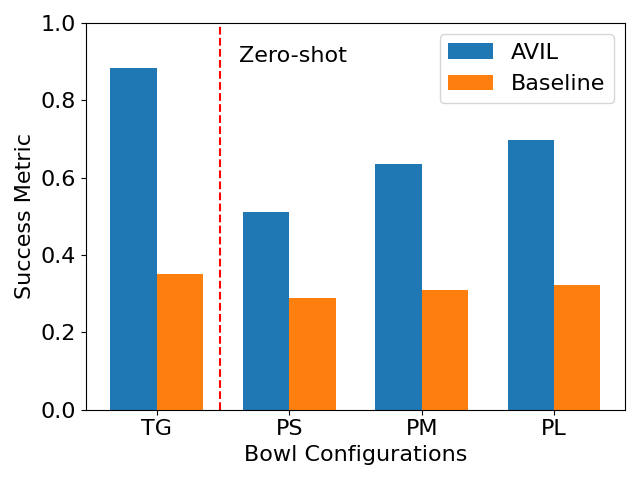}
    \caption{Performance comparison between \textbf{AVIL} and baseline across varied bowl configurations. We average the results over food types and bowl positions. We represent on the right side of the dashed red line conditions tested zero-shot.}
    \label{fig:exp_bowl}
\end{figure}

\begin{figure}[ht!]
    \centering
    \includegraphics[width=0.8\linewidth]{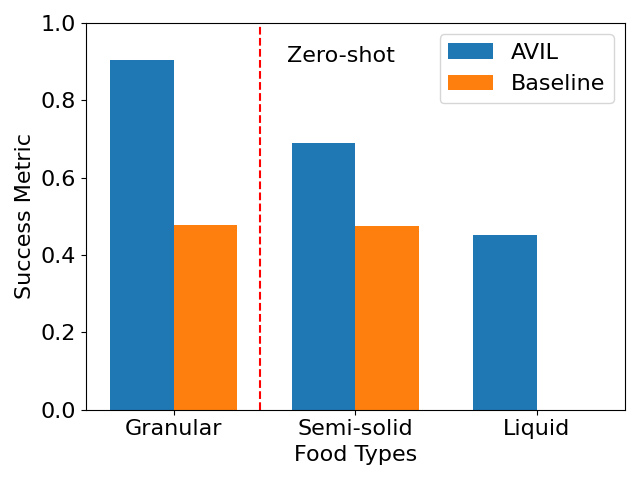}
    \caption{Performance comparison between \textbf{AVIL} and baseline across varied food types. We average the results over bowl configuration and bowl positions. We represent on the right side of the dashed red line conditions tested zero-shot.}
    \label{fig:exp_food}
\end{figure}

\begin{figure}[ht!]
    \centering
    \includegraphics[width=0.8\linewidth]{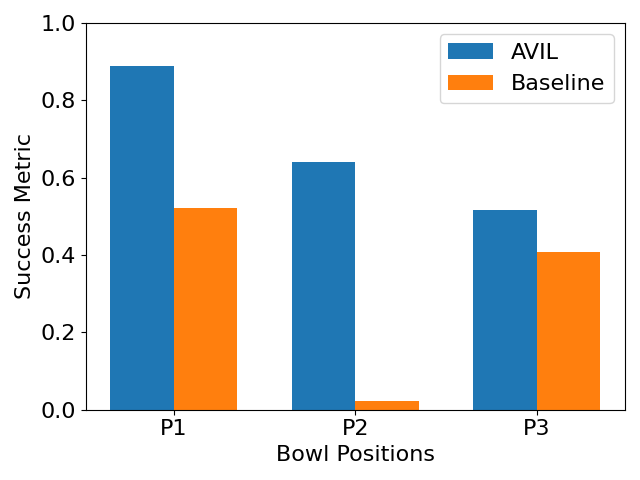}
    \caption{Performance comparison between \textbf{AVIL} and baseline across varied bowl positions. We average the results over bowl configurations and food types.}
    \label{fig:exp_pos}
\end{figure}

\begin{figure}[ht!]
    \centering
    \includegraphics[width=0.8\linewidth]{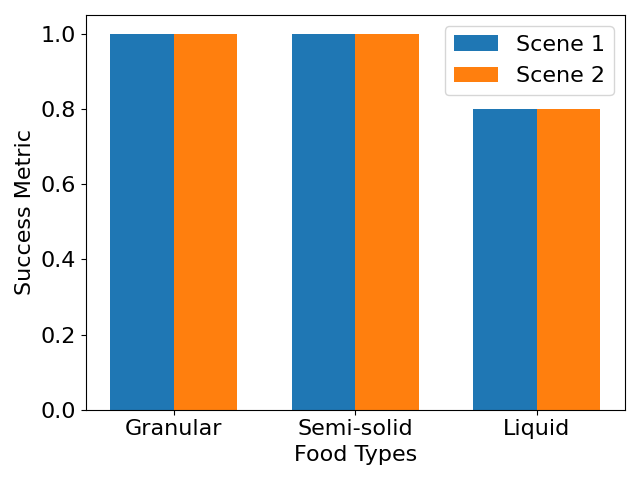}
    \caption{Performance comparison of \textbf{AVIL} with and without distractors with the TG bowl at position P1. For each food type, we conduct five trials of scooping attempts. Scene 1 represents without distractors and Scene 2 represents with distractors.}
    \label{fig:exp_scene}
\end{figure}

\subsection{Data Collection} \label{sec:data_collection}
We collected data through kinesthetic teaching, where a human operator maneuvered the robot arm to imitate scooping actions. Specifically, we utilized the TG bowl containing granular cereals, as depicted in Figure \ref{fig:pipeline}. The data collection process involved recording RGB images and robot joint positions throughout the scooping process. Notably, we did not collect data with plastic bowls of varying sizes or other types of food.

\subsection{Baseline}
For the baseline, we first utilize RetinaNet \cite{lin2017focal} to detect the bowl given an RGB image. Upon obtaining the bounding box, we calculate the centroid of the bowl. Subsequently, we map this position to the robot coordinate and direct the robot to move to that position with a specific height and orientation. Then, we adjust the wrist 2 joint of the robot arm to rotate by $-0.6$ radians to initiate the scooping action.

During testing on different bowl positions, the baseline maintains a consistent end-effector height and orientation to reach the bowl centroid. Additionally, the rotation of the wrist 2 joint remains fixed at $-0.6$ radians. We do not customize different baseline settings for varied bowl configurations and food types.

\subsection{Experimental Results}
In this section, we present and analyze the experimental results. Following the training of the visual imitation network, we evaluate its performance on a real robot. For illustrative purposes, in Figure \ref{fig:seq}, we showcase an example of our approach \textbf{AVIL} successfully scooping granular cereals from the TG bowl. We depict the motion of UR5e scooping cereals from left to right. 

To validate our approach \textbf{AVIL}, we also compare it with the baseline method. For both \textbf{AVIL} and basline, we test across varied bowl configurations, food types, and different positions as described in Section \ref{sec:exp_setup}. For each bowl configuration, food type, and position, we conduct five trials of scooping attempts. We use a success metric criterion. We assign a numerical value of 1 to successfully scooping food items from a bowl without spillage. We consider instances where some spillage occurs as partial success and assign a numerical value of 0.7. And we assign a numerical value of 0 to failure cases.

\subsubsection{\textbf{AVIL} and Baseline Performance Comparison}
We evaluate the performance of our approach \textbf{AVIL} and compare it with the baseline. To provide comprehensive comparisons, we average the success metrics over different aspects: when comparing success metrics across varied bowl configurations, we average over food types and bowl positions; likewise, when comparing success metrics across different food types, we average over bowl configurations and positions; and when comparing success metrics across bowl positions, we average over bowl configurations and food types.

\paragraph{Across varied bowl configurations}
In Figure \ref{fig:exp_bowl}, we illustrate the comparison results across varied bowl configurations, encompassing different bowl materials and sizes. The results demonstrate that \textbf{AVIL} consistently outperforms the baseline. Specifically, the success metric is 2.5, 1.8, 2.1, and 2.2 times higher compared to that of the baseline for TG, PS, PM, and PL bowls, respectively. The TG bowl achieves the highest success metric, as we collected the training data using this bowl. Additionally, among plastic bowls, the PS bowl has the lowest success metric, likely due to its smaller size, which increases the likelihood of spillage. 

\paragraph{Across varied food types}
In Figure \ref{fig:exp_food}, we present the comparison results across varied food types, including granular, semi-solid and liquid. Notably, the baseline struggles particularly with scooping liquid water due to its inherent property of flowing away. The handcrafted scooping motion employed by the baseline, involving rotation of the wrist 2 joint of the robot arm, proves insufficient in handling this challenge. In contrast, our approach adapts to such challenges by learning from human demonstrations, enabling effective scooping even with liquids. For granular and semi-solid food types, the success metric of \textbf{AVIL} is 1.7 and 1.3 times higher compared to that of the baseline, respectively. 

\paragraph{Across varied bowl positions}
In Figure \ref{fig:exp_pos}, we show the comparison results across varied bowl positions. For a fair comparison, \textbf{AVIL} and baseline have the same robot starting configuration for each trail of scooping attempts. The baseline faces difficulties with position P2, as it only directs the robot to move to the centroid of the bowl without planning a path in between. Consequently, the spoon collides with the bowl and fails to enter it during scooping attempts. However, our approach overcomes this limitation by learning from human demonstrations and effectively navigating the spoon into the bowl. For P1 and P3, the success metric of \textbf{AVIL} is 1.9 and 1.5 times higher compared to that of the baseline, respectively. 

\subsubsection{Zero-shot Generalization}
As detailed in Section \ref{sec:data_collection}, our data collection process exclusively involved the transparent glass bowl containing granular cereals. However, during testing, we evaluated \textbf{AVIL} on plastic bowls of various sizes and containing different food types such as liquid and semi-solid. Remarkably, \textbf{AVIL} demonstrates effective performance across these varied plastic bowls, as depicted in Figure \ref{fig:exp_bowl} on the right side of the dashed red line, with higher success metrics compared to the baseline for PS, PM, and PL bowls.

In Figure \ref{fig:exp_food}, positioned to the right of the dashed red line, when testing on semi-solid and liquid food types, VILA also exhibits effective performance. Specifically, the success metric for semi-solid is higher than that for liquid, which aligns with the expectation that liquid is more likely to flow away. Meanwhile, the baseline method proves ineffective for scooping liquid due to its inadequate motion.

\subsubsection{Robustness Against Distractors}
In the above Figure \ref{fig:att_map}, we demonstrate the effectiveness of our spatial attention module and its robustness against distractors. Here in Figure \ref{fig:exp_scene}, we present a performance comparison of \textbf{AVIL} with and without distractors. We denote Scene 1 as the condition without distractors and Scene 2 as the condition with distractors. We conduct tests on VILA using the TG bowl at position P1 with various food types, performing five trials of scooping attempts for each food type. As depicted in Figure \ref{fig:exp_scene}, both Scene 1 and Scene 2 exhibit identical performance. This suggests that distractors do not influence the performance of \textbf{AVIL}, verifying the robustness of \textbf{AVIL} against distractors.


\section{Conclusions} \label{sec:conclusion}
We introduce a novel visual imitation network with a spatial attention module for spoon scooping in RAF. Our approach, named AVIL (adaptive visual imitation learning), demonstrates adaptability and robustness, effectively handling varied bowl configurations in terms of material, size, and position, as well as diverse food types including granular, semi-solid, and liquid, even in the presence of distractors. This overcomes the drawbacks of previous work with limited adaptability to different container configurations and food types. To validate our approach, we conduct comprehensive experiments on a real robot and compare its performance with a baseline. The results demonstrate clear improvement over baseline across all variations,  with an enhancement of up to 2.5 times in terms of a success metric, validating the efficacy of our model.

\bibliographystyle{ieeetr}
\bibliography{references}
\end{document}